\documentclass[journal]{IEEEtran}

\usepackage{cite}
\usepackage{graphicx}
\usepackage{csquotes}
\usepackage{amsmath,amssymb,amsfonts}
\usepackage{amsthm}
\newtheorem{definition}{Definition}

\newtheorem*{prop}{Proposition}
\theoremstyle{definition}
\newtheorem{exmp}{Example}[section]
\usepackage{booktabs} 
\usepackage{float}
\newcommand{\ra}[1]{\renewcommand{\arraystretch}{#1}}

\def\BibTeX{{\rm B\kern-.05em{\sc i\kern-.025em b}\kern-.08em
    T\kern-.1667em\lower.7ex\hbox{E}\kern-.125emX}}

\begin{document}

\title{A Systems Theory of Transfer Learning}

\author{\IEEEauthorblockN{Tyler Cody and Peter A. Beling} \\
\IEEEauthorblockA{\textit{Engineering Systems and Environment} \\
\textit{University of Virginia}\\
Charlottesville, USA \\
}
}

\maketitle

\begin{abstract}

Existing frameworks for transfer learning are incomplete from a systems theoretic perspective. They place emphasis on notions of domain and task, and neglect notions of structure and behavior. In doing so, they limit the extent to which formalism can be carried through into the elaboration of their frameworks. Herein, we use Mesarovician systems theory to define transfer learning as a relation on sets and subsequently characterize the general nature of transfer learning as a mathematical construct. We interpret existing frameworks in terms of ours and go beyond existing frameworks to define notions of transferability, transfer roughness, and transfer distance. Importantly, despite its formalism, our framework avoids the detailed mathematics of learning theory or machine learning solution methods without excluding their consideration. As such, we provide a formal, general systems framework for modeling transfer learning that offers a rigorous foundation for system design and analysis.

\end{abstract}


\section*{Preliminaries}

Let $X$ denote a set and $x \in X$ denote its elements. For notational convenience random variables are not distinguished---probability measures on $X$ are denoted $P(X)$. The Cartesian product is denoted $\times$, and for any object $V_i = V_{i1} \times \ldots \times V_{in}$, $\overline{V_i}$ shall denote the family of component sets of $V_i$, $\overline{V_i} = \{ V_{i1}, \ldots, V_{in} \}$. The cardinality of $X$ is denoted $|X|$. The powerset is denoted $\mathcal{P}$. Herein it has two uses. Frequently, in order to express input-output conditions for a learning system we will only use its input-output representation $S:D \times X \to Y$. In contexts where $S \subset \times \{A, D, \Theta, H, X, Y\}$, we use $(d, x, y) \in \mathcal{P}(S)$ to make reference to the input-output representation. Also, the subset of the powerset of a powerset $K \subset \mathcal{P}(\mathcal{P}(D \cup \Theta))$ is used to denote that $K$ can be $\subset D$, $\subset \Theta$, or $\subset D \times \Theta$, etc., i.e., to make reference to ordered pairs. Often, we make reference to $d \in D$ to say a particular set of data $d$ from the larger set $D$. Additionally, for a system $S \subset X \times Y$, when we discuss $x \in X$ or $y \in Y$ it is assumed that $(x, y) \in S$ unless stated otherwise. This is to save the reader from the pedantry of Mesarovician abstract systems theory.

\section{Introduction}

Transfer learning, unlike classical learning, does not assume that the training and operating environments are the same, and, as such, is fundamental to the development of real-world learning systems. In transfer learning, knowledge from various \emph{source} sample spaces and associated probability distributions is \emph{transferred} to a particular \emph{target} sample space and probability distribution. Transfer learning enables learning in environments where data is limited. Perhaps more importantly, it allows learning systems to propagate their knowledge forward through distributional changes.

Mechanisms for knowledge transfer are a bottleneck in the deployment of learning systems. Learning in identically distributed settings has been the focus of learning theory and machine learning research for decades, however, such settings represent a minority of use cases. In real-world settings, distributions and sample spaces vary between systems and evolve over time. Transfer learning addresses such differences by sharing knowledge between learning systems, thus offering a theory principally based on distributional difference, and thereby a path towards the majority of use cases.

Existing transfer learning frameworks are incomplete from a systems theoretic perspective. They focus on domain and task, and neglect perspectives offered by explicitly considering system structure and behavior. Mesarovician systems theory can be used as a super-structure for learning to top-down model transfer learning, and although existing transfer learning frameworks may better reflect and classify the literature, the resulting systems theoretic framework offers a more rigorous foundation better suited for system design and analysis.

Mesarovician systems theory is a set-theoretic meta-theory concerned with the characterization and categorization of systems. A system is defined as a relation on sets and mathematical structure is sequentially added to those sets, their elements, or the relation among them to formalize phenomena of interest. By taking a top-down, systems approach to framing transfer learning, instead of using a bottom-up survey of the field, we naturally arrive at a framework for modeling transfer learning without necessarily referencing solution methods. This allows for general considerations of transfer learning systems, and is fundamental to the understanding of transfer learning as a mathematical construct.

We provide a novel definition of transfer learning systems, dichotomize transfer learning in terms of structure and behavior, and formalize notions of negative transfer, transferability, transfer distance, and transfer roughness in subsequent elaborations. First we review transfer learning and Mesarovician abstract systems theory in Section 2. We then define learning systems and discuss their relationship to abstract systems theory and empirical risk minimization in Section 3. Using this definition, transfer learning systems are defined and studied in Sections 4 and 5. We conclude with a synopsis and remarks in Section 6.

\begin{figure*}[t]
\centering
\includegraphics[width=1.0\textwidth]{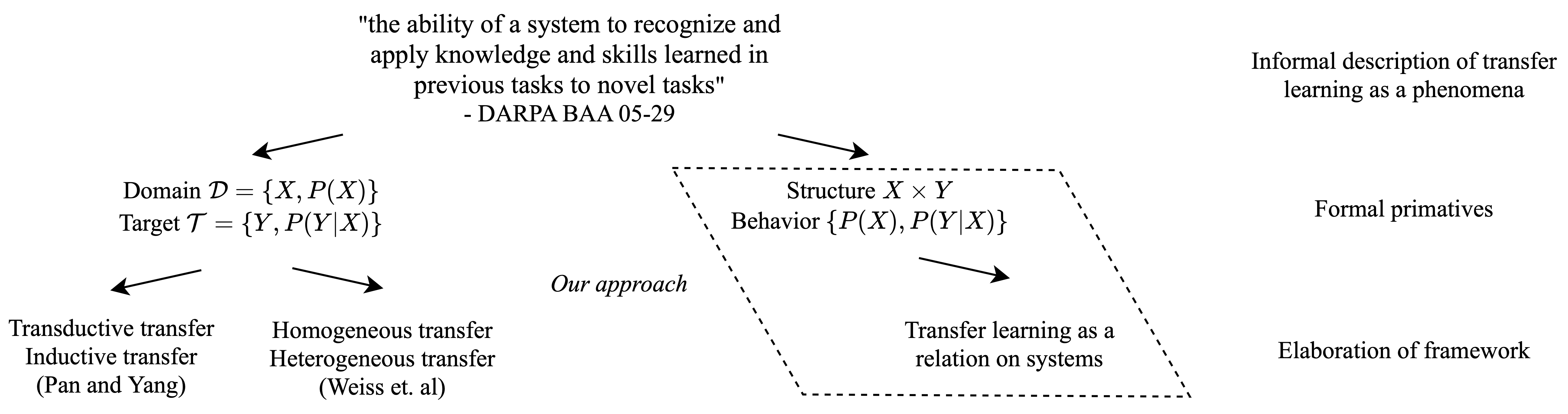}
\caption{Existing frameworks interpret the informal definition of transfer learning given by DARPA in terms of domain $\mathcal{D}$ and task $\mathcal{T}$. In contrast, we use structure and behavior, which provide a more formal basis for elaboration.}
\label{fig:frameworks}
\end{figure*}

\section{Background}

In the following we review transfer learning and make explicit the principal differences between existing frameworks and ours. Then, pertinent Mesarovician abstract systems theory is introduced. A supplemental glossary of Mesarovician terms can be found in the Appendix.

\subsection{Transfer Learning}

DARPA describes transfer learning as ``the ability of a system to recognize and apply knowledge and skills learned in previous tasks to novel tasks'' in Broad Agency Announcement (BAA) 05-29. The previous tasks are referred to as source tasks and the novel task is referred to as the target task. Thus, transfer learning seeks to transfer knowledge from some source learning systems to a target learning system.

Existing frameworks focus on a dichotomy between \emph{domain} $\mathcal{D}$ and \emph{task} $\mathcal{T}$. The domain $\mathcal{D}$ consists of the input space $X$ and its marginal distribution $P(X)$. The task $\mathcal{T}$ consists of the output space $Y$ and its posterior distribution $P(Y|X)$. The seminal transfer learning survey frames transfer learning in terms of an inequality of domains $\mathcal{D}$ and tasks $\mathcal{T}$ \cite{pan2009survey}. Therein, Pan and Yang define transfer learning as follows.
\begin{definition}{\emph{Transfer learning}.} \\
    Given a source domain $\mathcal{D}_S$ and task $\mathcal{T}_S$ and a target domain $\mathcal{D}_T$ and task $\mathcal{T}_T$, transfer learning aims to improve the learning of $P(Y_T|X_T)$ in the target using knowledge in $\mathcal{D}_S$ and $\mathcal{T}_S$, where $\mathcal{D}_S \neq \mathcal{D}_T$ or $\mathcal{T}_S \neq \mathcal{T}_T$.
\end{definition}

Pan and Yang continue by defining \emph{inductive transfer} as the case where the source and target tasks are not equal, $\mathcal{T}_S \neq \mathcal{T}_T$, and \emph{transductive transfer} as the case where the source and target domains are not equal but their tasks are, $\mathcal{D}_S \neq \mathcal{D}_T \land \mathcal{T}_S = \mathcal{T}_T$. They use these two notions, and their sub-classes, to categorize the transfer learning literature and its affinity for related fields of study. Alternative frameworks use notions of \emph{homogeneous} and \emph{heterogeneous} transfer, which correspond to the cases where the sample spaces of the source and target domains $X$ and tasks $Y$ are or are not equal, respectively \cite{weiss2016survey}.

While these formalisms describe the literature well, they are not rich enough to maintain formalism in the elaboration of their respective frameworks. For example, Pan and Yang address what, how, and when to transfer in a largely informal manner, making reference to inductive and transductive transfer as guideposts, but ultimately resorting to verbal descriptions \cite{pan2009survey}. In contrast, instead of starting with domain $\mathcal{D}$ and task $\mathcal{T}$ as the fundamental notions of transfer learning, we use structure and behavior---two concepts with deep general systems meaning, define transfer learning as a relation on systems, and carry formalism through into subsequent elaboration. The principal difference between existing frameworks and ours is depicted in Figure \ref{fig:frameworks}.

Importantly, despite our formalism, we maintain a general systems level of abstraction, in contrast to purely learning theoretical frameworks for transfer learning \cite{kuzborskij2013stability}. As such, we compare our general framework with those of Pan and Yang \cite{pan2009survey} and Weiss et. al \cite{weiss2016survey}. We greatly expand on previous, initial efforts in this direction \cite{cody2019systems, cody2020motivating}.

\begin{figure}[b]
\centering
\includegraphics[width=8cm]{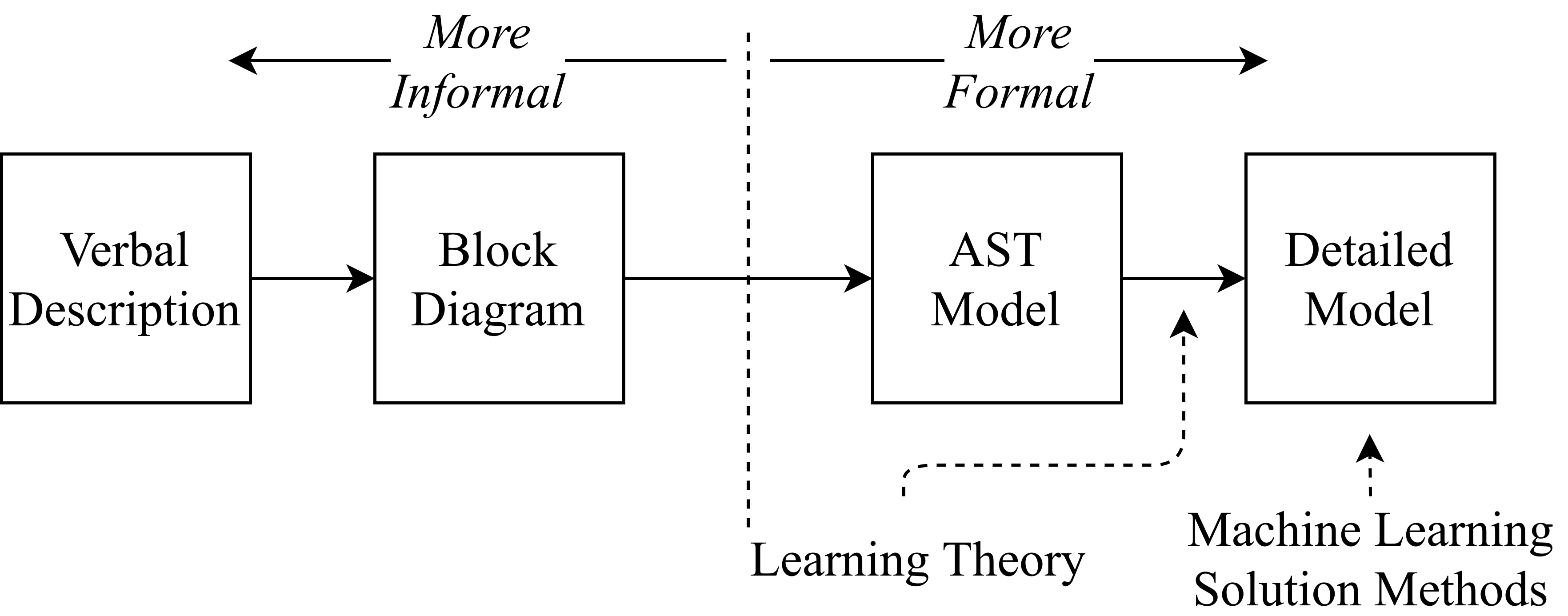}
\caption{AST is a minimally formal framework. In modeling learning, learning theory brings formalism to AST, and machine learning specifies the detailed model.}
\label{fig:ast-block}
\end{figure}

\subsection{Abstract Systems Theory}

Mesarovician abstract systems theory (AST) is a general systems theory that adopts the formal minimalist world-view \cite{mesarovic1989abstract, dori2019system}. AST is developed top-down, with the goal of giving a verbal description a parsimonious yet precise mathematical definition. Mathematical structure is added as needed to specify systems properties of interest. This facilitates working at multiple levels of abstraction within the same framework, where mathematical specifications can be added without restructuring the framework. In modeling, it is used as an intermediate step between informal reasoning and detailed mathematics by formalizing block-diagrams with little to no loss of generality, see Figure \ref{fig:ast-block}. Apparently this generality limits its deductive powers, but, in return, it helps uncover fundamental mathematical structure related to the general characterization and categorization of phenomena.

We will now review the AST definitions of a system, input-output system, and goal-seeking system, and the related notions of system structure and behavior. Additional details can be found in the Appendix.

In AST, a system is defined as a relation on component sets. When those sets can be partitioned, the system is called an input-output system. Systems and input-output systems are defined as follows.

\begin{definition}{\emph{System}.} \\
    A (general) system is a relation on non-empty (abstract) sets,
    $$S \subset \times \{ V_{i} : i \in I \}$$
    where $\times$ denotes the Cartesian product and $I$ is the index set. A component set $V_{i}$ is referred to as a system object.
\end{definition}

\begin{definition}{\emph{Input-Output Systems}.} \\
    Consider a system $S$, where $S \subset \times \{ V_{i} : i \in I \}$. Let $I_{x} \subset I$ and $I_{y} \subset I$ be a partition of $I$, i.e., $I_{x} \cap I_{y} = \emptyset$, $I_{x} \cup I_{y} = I$. The set $X = \times \{ V_{i} : i \in I_{x} \}$ is termed the input object and $Y = \times \{ V_{i} : i \in I_{y} \}$ is termed the output object. The system is then
    $$S \subset X \times Y$$
    and is referred to as an input-output system. If $S$ is a function $S:X \to Y$, it is referred to as a function-type system.
\end{definition}

\begin{figure}[t]
\centering
\includegraphics[]{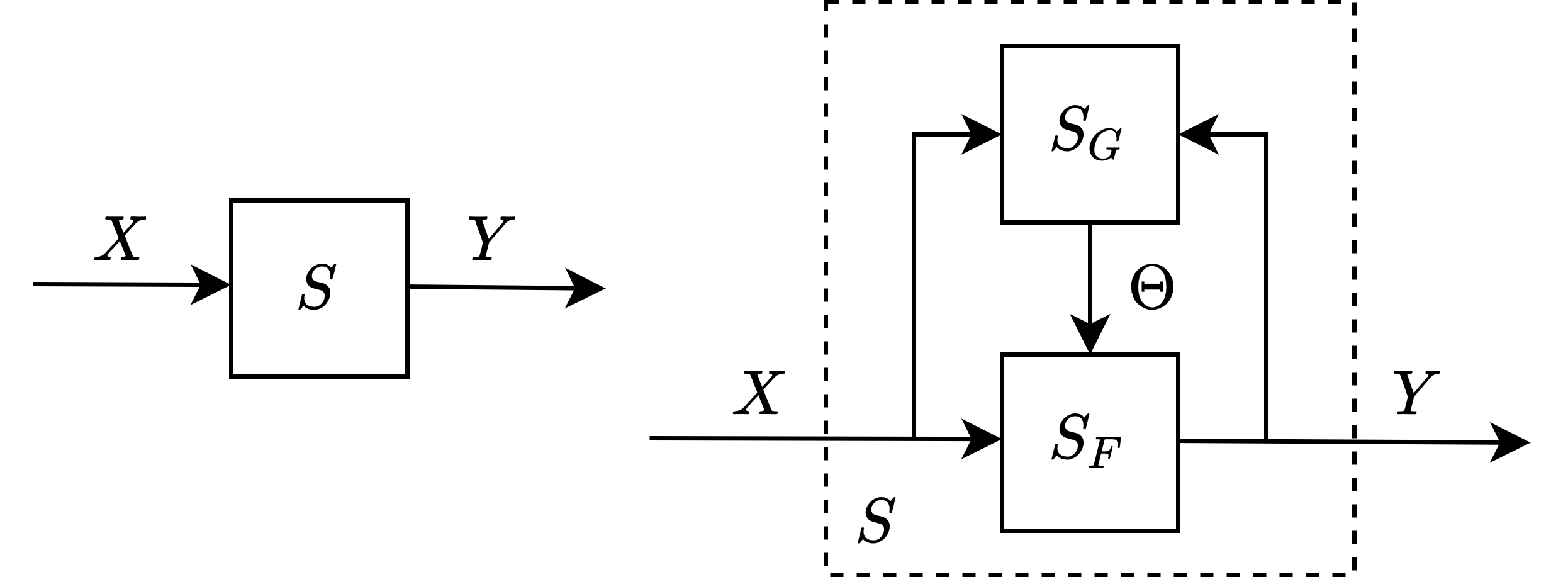}
\caption{Input-output systems (left) and goal-seeking systems (right).}
\label{fig:i-o-system}
\end{figure}

AST is developed by adding structure to the component sets and the relation among them. Input-output systems with an internal feedback mechanism are referred to as goal-seeking (or cybernetic) systems. The internal feedback of goal-seeking systems is specified by a pair of consistency relations $G$ and $E$ which formalize the notions of goal and seeking, respectively. Figure \ref{fig:i-o-system} depicts input-output and goal-seeking systems. Goal-seeking systems are defined as follows.

\begin{definition}{\emph{Goal-Seeking Systems}.} \\
    A system $S:X \to Y$ has a goal-seeking representation if there exists a pair of maps
    \begin{gather*}
      S_G:X \times Y \to \Theta \\
      S_F:\Theta \times X \to Y
    \end{gather*}
    and another pair
    \begin{gather*}
        G: \Theta \times X \times Y \to V \\
        E: X \times Y \times V \to \Theta
    \end{gather*}
    such that
    \begin{gather*}
        (x, y) \in S \leftrightarrow (\exists \theta) [(\theta, x, y) \in S_F \wedge (x, y, \theta) \in S_G] \\
        (x, y, G(\theta, x, y), \theta) \in E \leftrightarrow (x, y, \theta) \in S_G
    \end{gather*}
    where
    $$x \in X, y \in Y, \theta \in \Theta.$$
    $S_G$ is termed the goal-seeking system and $S_F$ the functional system. $G$ and $E$ are termed the goal and seeking relations, and $V$ the value.
    \label{def:gs}
\end{definition}

System structure and behavior are focal in Mesarovician characterizations of systems. System structure refers to the mathematical structure of a system's component sets and the relations among them. For example, there may be algebraic structure related to the specification of the relation, e.g. the linearity of a relationship between two component sets. System behaviors, in contrast, are properties or descriptions paired with systems. For example, consider a system $S:X \to Y$ and a map $S \to \{ stable, neutral, unstable \}$. A linear increasing function and an increasing power function may both be considered behaviorally unstable, but clearly their structures are different \cite{mesarovic1989abstract}.

Similarity of systems is a fundamental notion, and it can be expressed well in structural and behavioral terms. Structural similarity describes the \emph{homomorphism} between two systems' structures. Herein, in accord with category theory, a map from one system to another is termed a morphism, and homomorphism specifies the morphism to be onto. Homomorphism is formally defined as follows.
\begin{definition}{\emph{Homomorphism}.} \\
    An input-output system $S \subset X \times Y$ is homomorphic to $S' \subset \times X' \times Y'$ if there exists a pair of maps,
    \begin{align*}
        \varrho:X \to X', \vartheta:Y \to Y'
    \end{align*}
    such that for all $x\in X$, $x'\in X'$, and $y\in Y$, $y'\in Y'$, $\varrho(x)=x'$ and $\vartheta(y)=y'$.
\end{definition}
\noindent Behavioral similarity, in contrast, describes the \emph{proximity} or \emph{distance} between two systems' behavior. As in AST generally, we use structure and behavior as the primary apparatus for elaborating on our formulation of transfer learning systems. Refer to the Appendix for additional details on structure, behavior, and similarity.

\section{Learning Systems}

We follow Mesarovic's top-down process to sequentially construct a learning system $S$. Learning is a relation on data and hypotheses. To the extent that a scientific approach is taken, those hypotheses are explanations of initial-final condition pairs \cite{popper2005logic}. Otherwise put, we are concerned with learning as function estimation. We additionally note that learning algorithms use data to select those hypotheses and that the data is a sample of input-output pairs \cite{vapnik2013nature}. Such a learning system can be formally defined as follows.

\begin{definition}{\emph{(Input-Output) Learning System}.} \\
    A learning system $S$ is a relation
    $$S \subset \times \{A, D, \Theta, H, X, Y \}$$
    such that
    \begin{gather*}
        D \subset X \times Y, A:D \to \Theta, H:\Theta \times X \to Y \\
        (d, x, y) \in \mathcal{P}(S) \leftrightarrow (\exists \theta) [(\theta, x, y) \in H \wedge (d, \theta) \in A]
    \end{gather*}
    where 
    $$x \in X, y \in Y, d \in D, \theta \in \Theta.$$ 
    The algorithm $A$, data $D$, parameters $\Theta$, hypotheses $H$, input $X$, and output $Y$ are the component sets of $S$, and learning is specified in the relation among them.
    \label{def:ls}
\end{definition} 

The above definition of learning formalizes learning as a cascade connection of two input-output systems: an inductive system $S_I \subset \times \{ A, D, \Theta \}$ responsible for inducing hypotheses from data, and a functional system $S_F \subset \times \{ \Theta, H, X, Y \}$, i.e. the induced hypothesis. $S_I$ and $S_F$ are coupled by the parameter $\Theta$. Learning is hardly a purely input-output process, however. To address this, we must specify the goal-seeking nature of $S_I$, and, more particularly, of $A$.

$A$ is goal-seeking in that it makes use of a \emph{goal} relation $G: D \times \Theta \to V$ that assigns a value $v \in V$ to data-parameter pairs, and a \emph{seeking} relation $E: V \times D \to \Theta$ that assigns parameter $\theta \in \Theta$ to data-value pairs. These consistency relations $G$ and $E$ specify $A$, but not by decomposition; i.e., in general, $G$ and $E$ cannot be composed to form $A$. The definition of a learning system can be extended as follows.

\begin{definition}{\emph{(Goal-Seeking) Learning System}.} \\
    A learning system $S$ is a relation
    $$S \subset \times \{ A, D, \Theta, G, E, H, X, Y \}$$
    such that
    \begin{gather*}
        D \subset X \times Y, A:D \to \Theta, H:\Theta \times X \to Y \\
        (d, x, y) \in \mathcal{P}(S) \leftrightarrow (\exists \theta) [(\theta, x, y) \in H \wedge (d, \theta) \in A] \\
        G: D \times \Theta \to V, E: V \times D \to \Theta \\
        (d, G(\theta, d), \theta) \in E \leftrightarrow (d, \theta) \in A
    \end{gather*}
    where
    $$x \in X, y \in Y, d \in D, \theta \in \Theta.$$
    The algorithm $A$, data $D$, parameters $\Theta$, consistency relations $G$ and $E$, hypotheses $H$, input $X$, and output $Y$ are the component sets of $S$, and learning is specified in the relation among them.
    \label{def:gsls}
\end{definition} 

\begin{figure}[t]
\centering
\includegraphics[]{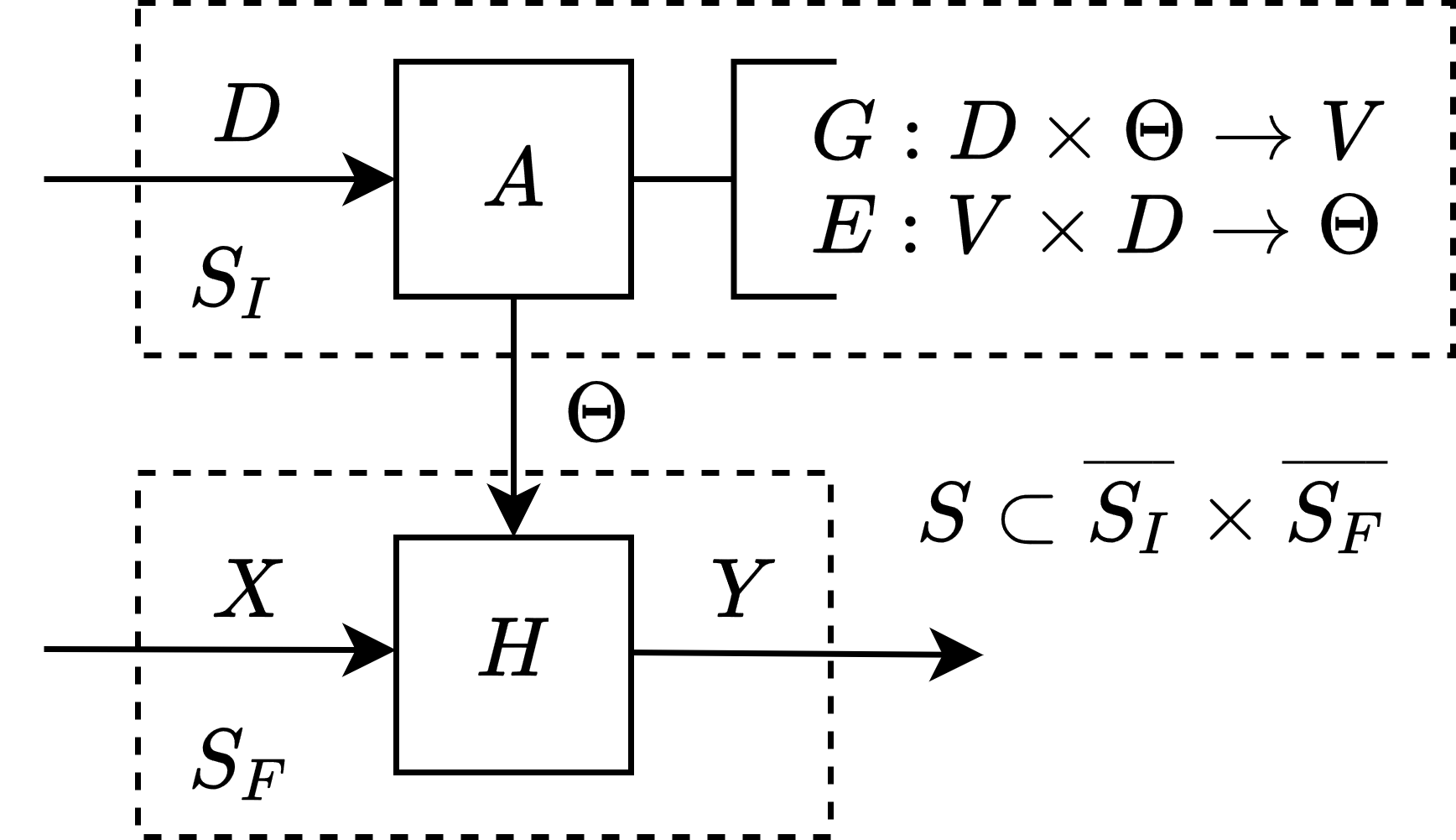}
\caption{Learning systems are a cascade connection of the inductive system $S_I$ and the induced hypothesis $S_F$. $S_I$ is goal-seeking.}
\label{fig:learning-systems}
\end{figure}

Learning systems are depicted in Figure \ref{fig:learning-systems}. These systems theoretic definitions of learning have an affinity to learning theoretic constructions. Consider empirical risk minimization (ERM), where empirical measures of risk are minimized to determine the optimal hypothesis for a given sample \cite{vapnik2013nature}. Apparently, ERM specifies $G$ to be a measure of risk calculated on the basis of a sample drawn independently according to a probability measure on the approximated function $f:X \to Y$ and specifies $E$ to be a minimization of $G$ over $\Theta$.

We have demonstrated how our definition of a learning system anchors our framework to both AST and ERM. We posit these definitions not as universal truths, but rather as constructions that anchor our framing of transfer learning to systems and learning theory. We abstain from further elaboration on these definitions, however, proofs of the above propositions can be found in the Appendix. In the following, we leave $G$ and $E$ implicit, only making reference to $f$ and related probability measures.

\begin{exmp}{\emph{Learning in an Unmanned Aerial Vehicle.}} \\
    Consider an unmanned aerial vehicle (UAV) with a learning system $S$ for path planning. $H$ is a function from sensor data $X$, e.g., from accelerometers, cameras, and radar, to flight paths $Y$. $D$, then, consists of sets of sensor-path pairs. If $S$ is a support-vector machine (SVM), then $H$ is a set of half-spaces parameterized by $\Theta$ and $A$ is a convex optimization routine\cite{suykens1999least}.
    \label{ex:learning-systems}
\end{exmp}

\section{Transfer Learning Systems}

Transfer learning is conventionally framed as a problem of sharing knowledge from source domains and tasks to a target domain and task. We propose an alternative approach. We formulate transfer learning top-down in reference to the source and target learning systems, and then dichotomize subsequent analysis not by domain and task, but rather by structure, described primarily by the $X \times Y$ space, and behavior, described primarily by probability measures on the estimated function $f:X \to Y$. 

A transfer learning system is a relation on the source and target systems that combines knowledge from the source with data from the target and uses the result to select a hypothesis that estimates the target learning task $f_T$. We define it formally as follows.

\begin{definition}{\emph{Transfer Learning System}.} \\
    Given source and target learning systems $S_S$ and $S_T$
    \begin{gather*}
        S_S \subset \times \{ A_S, D_S, \Theta_S, H_S, X_S, Y_S \} \\
        S_T \subset \times \{ A_T, D_T, \Theta_T, H_T, X_T, Y_T \} 
    \end{gather*}
    a transfer learning system $S_{Tr}$ is a relation on the component sets of the source and target systems $S_{Tr} \subset \overline{S_S} \times \overline{S_T}$ such that
    $$K_S \subset D_S \times \Theta_S, D \subset D_T \times K_S$$
    and
    \begin{gather*}
        A_{Tr}: D \to \Theta_{Tr}, H_{Tr}: \Theta_{Tr} \times X_T \to Y_T \\
        (d, x_T, y_T) \in \mathcal{P}(S_{Tr}) \leftrightarrow \\
        (\exists \theta_{Tr})[(\theta_{Tr}, x_T, y_T) \in H_{Tr} \land (d, \theta_{Tr}) \in A_{Tr}]
    \end{gather*}
    where
    $$x_T \in X_T, y_T \in Y_T, d \in D, \theta_{Tr} \in \Theta_{Tr}.$$
    The nature of source knowledge $K_S$\footnote{Here, we define the transferred knowledge $K_S$ to be $D_S$ and $\Theta_S$, the source data and parameters, following convention \cite{pan2009survey}. In general, however, source knowledge $K_S \subset \mathcal{P}(\mathcal{P}(\overline{S_S}))$.}, the transfer learning algorithm $A_{Tr}$, hypotheses $H_{Tr}$, and parameters $\Theta_{Tr}$ specify transfer learning as a relation on $\overline{S_S}$ and $\overline{S_T}$. 
    \label{def:tl}
\end{definition} 

Trivial transfer occurs when the structure and behavior of $S_S$ and $S_T$ are the same, or, otherwise put, when transfer learning reduces to classical, identically distributed learning. Transfer is non-trivial when there is a structural difference $X_S \times Y_S \neq X_T \times Y_T$ or a behavioral difference $P(X_S) \neq P(X_T) \lor P(Y_S|X_S) \neq P(Y_T|X_T)$ between the source $S_S$ and target $S_T$. If the posterior distributions $P(Y|X)$ and marginal distributions $P(X)$ are equal between the source and target systems, then transfer is trivial. Non-trivial transfer is implied when $X_S \times Y_S \neq X_T \times Y_T$.

\begin{prop}
    \emph{$S_{Tr}$ in Definition \ref{def:tl} is a learning system as defined in Definition \ref{def:ls}.}\\
    \emph{Proof:}
    As stated in Definition \ref{def:tl}, a transfer learning system is a relation $S_{Tr} \subset \overline{S_S} \times \overline{S_T}$. More particularly, it is a relation $S_{Tr} \subset (D_S \times \Theta_S) \times (D_T \times X_T \times Y_T)$, and has a function-type representation $S_{Tr}: D_S \times \Theta_S \times D_T \times X_T \to Y_T$. Its inductive system is the relation $A_{Tr}:D \to \Theta_{Tr}$, where $D \subset D_S \times \Theta_S \times D_T$. And its functional system is the relation $H_{Tr}: \Theta_{Tr} \times X_T \to Y_T$. Thus, we can restate $S_{Tr}$ as a relation
    $$S_{Tr} \subset \times \{A_{Tr}, D, \Theta_{Tr}, H_{Tr}, X_T, Y_T\}$$
    and since by Definition \ref{def:tl}
    \begin{gather*}
        (d, x_T, y_T) \in \mathcal{P}(S_{Tr}) \leftrightarrow \\
        (\exists \theta_{Tr})[(\theta_{Tr}, x_T, y_T) \in H_{Tr} \land (d, \theta_{Tr}) \in A_{Tr}]
    \end{gather*}
    where
    $$x_T \in X_T, y_T \in Y_T, d \in D, \theta_{Tr} \in \Theta_{Tr},$$
    we have that $S_{Tr}$ is an input-output learning system as in Definition $\ref{def:ls}$.
\end{prop}

\begin{figure}[t]
\centering
\includegraphics[]{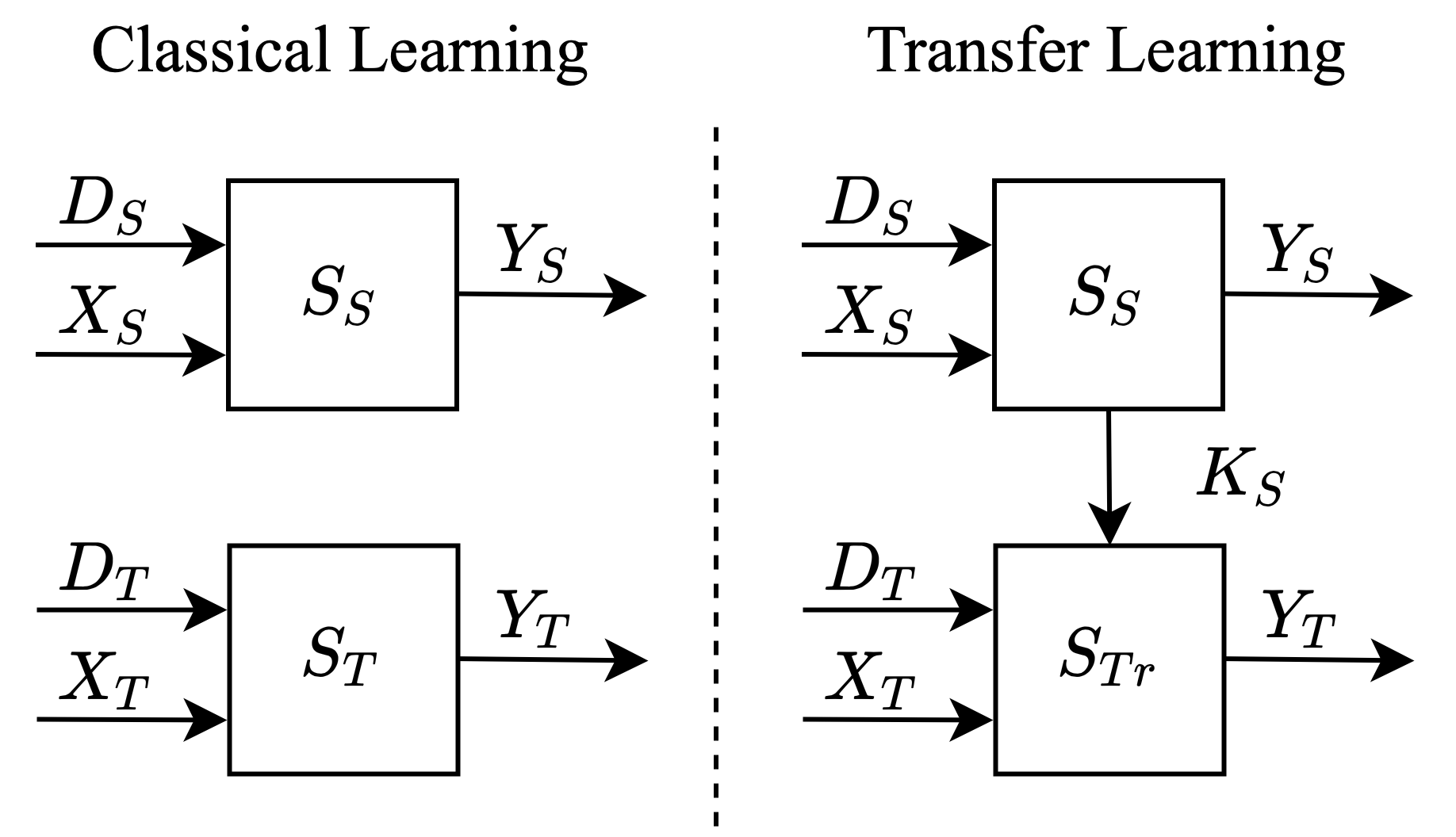}
\caption{Transfer learning systems $S_{Tr}$ are a relation $K_S \times D_T \times X_T \to Y_T$, while the target system $S_T$ is a relation $D_T \times X_T \to Y_T$.}
\label{fig:tl-system}
\end{figure}

Transfer learning systems are distinguished from general learning systems by the selection and transfer of $K_S$, and its relation to $D_T$ by way of $D \subset K_S \times D_T$ and its associated operator $K_S \times D_T \to D$. In cases where $\{A_{Tr}, \Theta_{Tr}, H_{Tr}\} \leftrightarrow \{A_T, \Theta_T, H_T\}$, e.g., as is possible when transfer learning consists of pooling samples with identical supports, the additional input $K_S$ is all that distinguishes $S_{Tr}$ from $S_T$. Classical and transfer learning systems are depicted in Figure \ref{fig:tl-system}.

As we will see, however, this is no small distinction, as it allows for consideration of learning across differing system structures and behaviors. But before we elaborate on the richness of structural and behavioral considerations, first, in the following subsections, we interpret existing frameworks in terms of structure and behavior and define preliminary notions related to generalization in transfer learning.

\begin{exmp}{\emph{Transfer Learning in UAVs.}} \\
    Consider UAVs with learning systems $S_S$ and $S_T$ defined according to Example \ref{ex:learning-systems} and a transfer learning system $S_{Tr} \subset \overline{S_S} \times \overline{S_T}$. If $S_{Tr}$ is also a SVM, then $H_{Tr}$ are also half-spaces parameterized by $\Theta_{Tr}$. If $K_S \subset D_S \times \Theta_S$, $\Theta_S$ can provide an initial estimate for $\Theta_{Tr}$, and $D_S$ can be pooled with $D_T$ to update this estimate. $A_{Tr}$, in distinction to $A_T$, must facilitate this initialization and pooling.
    \label{ex:transfer}
\end{exmp}

\subsection{Comparison to Existing Frameworks}

Using Definition \ref{def:tl}, the central notions of existing frameworks can be immediately defined in terms of structural and behavioral inequalities. Homogeneous transfer specifies structural equality of the source and target sample spaces, $X_S \times Y_S = X_T \times Y_T$, and heterogeneous transfer specifies otherwise. Domain adaptation, co-variate shift, and prior shift are all examples of homogeneous transfer \cite{jiang2008literature, pan2009survey, csurka2017comprehensive}. Transductive and inductive transfer entail more nuanced specifications. 

Recall, inductive transfer specifies that $\mathcal{T}_S \neq \mathcal{T}_T$ and transductive transfer specifies that $\mathcal{D}_S \neq \mathcal{D}_T \land \mathcal{T}_S = \mathcal{T}_T$, where $\mathcal{D}=\{P(X), X\}$ and $\mathcal{T}=\{P(Y|X), Y\}$. Technically, transductive transfer occurs if $X_S \neq X_T$ or if $P(X_S) \neq P(X_T)$. However, if $X_S \neq X_T$, then it is common for $P(Y_S|X_S) \neq P(Y_T|X_T)$ because the input set conditioning the posterior has changed, and thus it is likely that $\mathcal{T_S} \neq \mathcal{T_T}$. To that extent, in the main, transductive transfer specifies a difference between input behavior while output behavior remains equal. Inductive transfer, on the other hand, is more vague, and merely specifies that there is a structural difference in the outputs, $Y_S \neq Y_T$, or a behavioral difference in the posteriors, $P(Y_S|X_S) \neq P(Y_S|X_T)$. Note, this behavioral difference in the posteriors can be induced by a structural difference in the inputs as previously mentioned, and is implied by a structural difference in the outputs.

In short, the homogeneous-heterogeneous dichotomy neglects behavior and the transductive-inductive framing muddles the distinction between structure and behavior. While frameworks based on either cover the literature well, they only provide high-level formalisms which are difficult to carry through into general, formal characterizations of transfer learning systems. In contrast, Definition \ref{def:tl} provides a formalism that can be used to define transfer learning approaches and auxiliary topics in generalization.

\subsection{Transfer Approaches}

Consider how the seminal framework informally classifies transfer learning algorithms \cite{pan2009survey}. Three main approaches are identified: `instance transfer', `parameter transfer', and `feature-representation transfer'. While the transductive or inductive nature of a transfer learning system gives insight into which approaches are available, the approaches cannot be formalized in those terms, or in terms of domain $\mathcal{D}$ and task $\mathcal{T}$ for that matter, because they are a specification on the inductive system $S_I \subset \times \{A_{Tr}, D_{Tr}, \Theta_{Tr}\}$, whereas the former are specifications on the functional system $S_F \subset \times \{\Theta_{Tr}, H_{Tr}, X_{Tr}, Y_{Tr}\}$.

With the additional formalism of Definition \ref{def:tl}, these transfer approaches can be formalized using system structure. First, note that differently structured data $D$ leads to different approaches. Consider the categories of transfer learning systems corresponding to the various cases where $D \subset \mathcal{P}(\mathcal{P}(D_T \cup D_S \cup \Theta_S))$. Instance and parameter transfer correspond to transferring knowledge in terms of $D_S$ and $\Theta_S$, respectively, and can be formally defined as follows.
\begin{definition}{\emph{Instance Transfer}.} \\
    A transfer learning system $S_{Tr}$ is an instance transfer learning system if $K_S \subset D_S$, i.e., if 
    $$\mathcal{A}_{Tr}: D \to \Theta_{Tr} \iff \mathcal{A}_{Tr}: D_S \times D_T \to \Theta_{Tr}.$$
\end{definition}

\begin{definition}{\emph{Parameter Transfer}.} \\
    A transfer learning system $S_{Tr}$ is a parameter transfer learning system if $K_S \subset \Theta_S$, i.e., if 
    $$\mathcal{A}_{Tr}: D \to \Theta_{Tr} \iff \mathcal{A}_{Tr}: \Theta_S \times D_T \to \Theta_{Tr}.$$
\end{definition}

Feature-representation transfer, in contrast, specifies that learning involves transformations on $\overline{S_T}$, $K_S$, or both. It can be defined formally as follows. 

\begin{definition}{\emph{Feature-Representation Transfer}.} \\
    Consider a transfer learning system $S_{Tr}$ and a learning system $S_L$, termed the latent learning system. Note, $S_{Tr}$ and $S_L$ can be represented as function-type systems,
    \begin{gather*}
        S_{Tr}: D \times X_T \to Y_T \\
        S_L: D_L \times X_L \to Y_L.
    \end{gather*}
    $S_{Tr}$ is a feature-representation transfer learning system if there exist maps
    $$m_D:D \to D_L, m_{X_T}:X_T \to X_L, m_{Y_L}:Y_L \to Y_T$$
    such that
    \begin{gather*}
        \forall (d, x_T, y_T) \in (S_{Tr}) \\
        S_{Tr}(d, x_T) \leftrightarrow m_{Y_L}(S_L(m_D(d), m_{X_T}(x_T)))
    \end{gather*}
    where
    $$d \in D, x_T \in X_T, y_T \in Y_T.$$
    In other words, $S_{Tr}$ is a feature-representation transfer learning system if transfer learning involves transforming to and from a latent system where learning occurs.
\end{definition}

\begin{prop}
    \emph{Learning in $S_S$, $S_T$, and $S_L$.} \\
    Consider a case of feature-representation transfer where $K_S \subset D_S$. Let $m_{D_T}:D_T \to D_L$ and $m_{D_S}:D_S \to D_L$. Then, $m_D \iff (m_{D_T}, m_{D_S})$. Recall $D_i \subset X_i \times Y_i$. If $m_{D_T}$ is the identity and $m_{D_S}$ is not, then $X_T \times Y_T = X_L \times Y_L$---learning occurs in the target sample space. If $m_{D_S}$ is the identity and $m_{D_T}$ is not, then $X_S \times Y_S = X_L \times Y_L$---learning occurs in the source sample space. If $m_D$ is the identity, then $X_S \times Y_S = X_T \times Y_T = X_L \times Y_L$, i.e., $S_{Tr}$ involves homogeneous transfer. If neither $m_{D_T}$ or $m_{D_S}$ are the identity, then learning occurs in a latent sample space $X_L \times Y_L$ that is unequal to $X_T \times Y_T$ and $X_S \times Y_S$.
\end{prop}

In feature-representation transfer, data $D \subset D_T \times K_S$ is mapped to a latent system $S_L$ where learning occurs. By way of $m_D:D \to D_L$, feature-representation transfer involves relating the source and target input-output spaces to a latent space $X_L \times Y_L$. Learning can occur in $X_L \times Y_L$, and, using $m_{Y_L}$, the output can be given in terms of the target output $Y_T$. Similarly, the target can be mapped onto the source, $X_L \times Y_L = X_S \times Y_S$, where learning can occur given $m_{Y_L}$, or the source can be mapped onto the target, $X_L \times Y_L = X_T \times Y_T$. 

\begin{figure*}[t]
\centering
\includegraphics[]{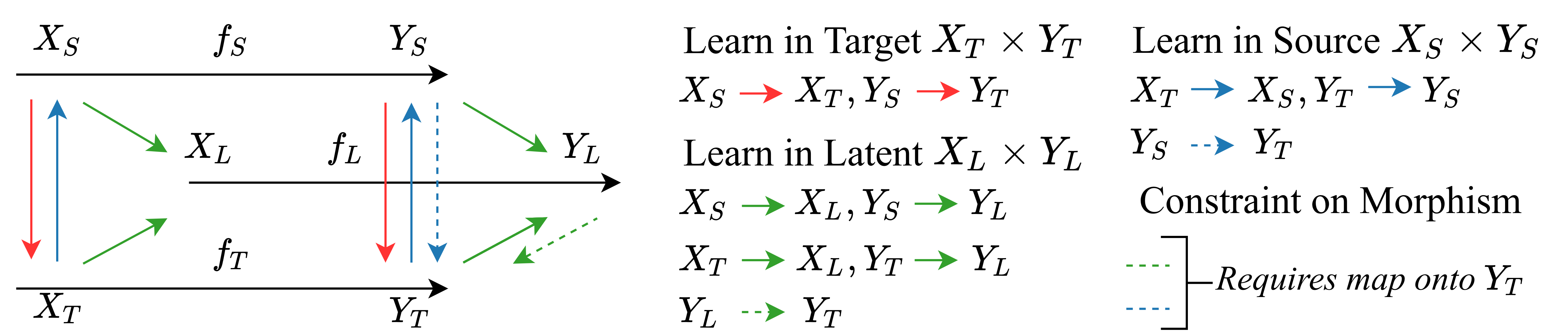}
\caption{Morphisms in feature representation learning. Learning in the target sample space requires a morphism from that of the source, as shown in red. Learning in the source sample space requires a morphism from that of the target, as shown in blue, and a map from the source output to the target output, as shown by the dashed blue arrow. And learning in a latent sample space requires morphisms from both the source and target sample spaces to that of the latent system, as shown in green, and a map from the latent output to the target output, as shown by the dashed green arrow. As discussed in Section 5, the nature of these morphisms affects the difficulty of transfer.}
\label{fig:latent-learning}
\end{figure*}

Figure \ref{fig:latent-learning} depicts these three cases of morphisms using a commutative diagram. As the individual maps that compose these morhpisms become more dislike identities and partial, feature-representation transfer becomes more difficult. We will discuss this further in our elaboration on structural considerations. Additionally note, even if $X_S \times Y_S = X_T \times Y_T$, feature-representation transfer may still be used to better relate source and target behavior. 

\begin{table}[t]
\centering
\ra{1.3}
\begin{tabular}{@{}ll@{}}
\toprule
Transfer Approach & Algorithm Structure\\
\midrule
Instance & $A_{Tr}:D_T \times D_S \to \Theta_{Tr}$\\
Parameter & $A_{Tr}:D_T\ \times \Theta_S \to \Theta_{Tr}$\\
Instance \& Parameter & $A_{Tr}:D_T \times D_S \times \Theta_S \to \Theta_{Tr}$\\
Feature-Representation & $A_{Tr}: m_D(D) \to \Theta_{Tr}$ \\
\bottomrule
\\
\end{tabular}
\caption{Structural differences between transfer approaches.}
\label{table:approaches}
\end{table}

Instance, parameter, and feature-based approaches are shown in terms of their specification on transfer learning algorithms $A_{Tr}$ in Table \ref{table:approaches}. Another general notion in transfer learning is \emph{n}-shot transfer. It can be defined as follows.
\begin{definition}{\emph{N-shot Transfer}.} \\
    A transfer learning system $S_{Tr}$ with target data $d_T \in D_T$ is referred to as a n-shot transfer learning system if $|d_T| = n$. Zero-shot transfer occurs if $A_{Tr}: D \to \Theta_{Tr} \iff A_{Tr}: K_S \to \Theta_{Tr}$.
\end{definition}
\noindent Machine learning is often concerned with few-shot learners---transfer learning systems that can generalize with only a few samples from the target. We will discuss generalization in transfer learning in the following subsection, but first, to get a sense of how we formalize instance, parameter, and feature-representation transfer, consider how a few canonical transfer learning algorithms are modeled by our framework.

Transfer component analysis uses a modified principal component analysis approach to project the source and target data into a relatable latent space \cite{pan2010domain}, i.e., it is an instance approach in that $D_S$ is used in $A_{Tr}$ and a feature-representation approach in that $X_S$ and $X_T$ are projected into a latent $X_L$. Constraining parameters to be within a range of those of the source, as in hierarchical Bayesian and regularization approaches, is parameter transfer \cite{evgeniou2004regularized, schwaighofer2005learning}. Deep learning approaches often involve parameter transfer in that the weights $\Theta_S$ of the source network are shared and frozen in the target, or otherwise used to initialize $\Theta_T$ \cite{bengio2012deep}. Other deep learning approaches also involve instance transfer to increase sample size, such as those that use generative adversarial networks \cite{sankaranarayanan2018generate}. When the source and target data must first be transformed before the data can be related, they are also feature-representation approaches, as in joint adaptation networks \cite{long2017deep}.

By formalizing the canonical classes of transfer approaches, we are better able to understand them in terms of their general requirements on $S_{Tr}$, particularly on $S_I$, and more particularly on $A_{Tr}$ and $D$. The informal use of these classes by existing frameworks, wherein a solution method's dominant nature sorts it into a particular class, does well to organize the literature. Our formalisms can cloud these scholarly distinctions, as shown in the case of deep learning where a single method can belong to all three classes, however, they give a basis for defining formal categories of transfer learning systems $S_{Tr}$ in terms of their inductive systems $S_I$. 

\subsection{Generalization in Transfer Learning}

Generalization is, perhaps, the ultimate aim of learning.  It is the ability for the learned hypothesis to approximate $f$ out-of-sample, i.e., on samples not seen in training. Generalization as a goal for learning systems is implicit in $A$ when a measure of error $\epsilon$ between $h(\theta)$ and $f$ specifies $G$, such as in ERM. Herein, we define it as follows.
\begin{definition}{\emph{Generalization.}} \\
Given a learning system $S$ and data $d \in D$, generalization is the ability for a learned hypothesis $h(\theta)$ to estimate learning task $f:X \to Y$, on samples $(x, y) \notin d$.
\end{definition}

In moving from the classical, identically distributed learning setting to transfer learning, we move from generalizing to a new sample from the same system, to generalizing to a new sample from a different system. In classical learning, for a learning system $S$, the estimated function $f$ is specified by $P(Y|X)$ and data $D$ are drawn from a related joint $P(X,Y)$. In transfer learning, however, the $X \times Y$ space and probability measures specifying $f$ and $D$ vary between $S_S$ and $S_T$. 

In classical learning, given a learning system $S$, data $d \in D$, a measure of error $\epsilon: H(\Theta) \times f \to \mathbb{R}$, and a threshold on error $\epsilon^* \in \mathbb{R}$, we generalize if 
$$\epsilon(H(A(d)), f) \leq \epsilon^*.$$
That, is, if the measure of error between the learned hypothesis and the function it estimates is below a threshold. In practice, since $f$ is not known, error is empirically estimated using samples $(x, y) \in X \times Y$ such that $(x, y) \notin d$. 

In transfer learning, given $S_{Tr}$ and data $d \in D$, we generalize if 
$$\underbrace{\epsilon(H_{Tr}(A_{Tr}(d)), f_T)}_{\epsilon_T} \leq \epsilon^*.$$
If $\epsilon_T$ is smaller without any transferred knowledge from $S_S$ than with, transfer from $S_S$ to $S_T$ is said to result in negative transfer. Negative transfer is defined in accord with Wang et. al as follows.
\begin{definition}{\emph{Negative Transfer}.} \\
    Consider a transfer learning system $S_{Tr}$. Recall $D \subset D_T \times K_S$. Let $d \in D$ and $d_T \in D_T$. Given a measure of error $\epsilon: H(\Theta) \times f \to \mathbb{R}$, negative transfer is said to occur if
    $$\epsilon(H_T(A_T(d_T)), f_T) < \epsilon(H_{Tr}(A_{Tr}(d)), f_T),$$
    that is, if the error in estimating $f_T$ is higher with the transferred knowledge than without it.
    \label{def:nt}
\end{definition}
\noindent As Wang et. al note, negative transfer can arise from behavioral dissimilarity between the source and target \cite{wang2019characterizing}. In general, it can arise from structural dissimilarity as well.

Because generalization in transfer learning considers generalization across systems, as opposed to generalization within a given system, naturally, it is concerned with the set of systems to and from which transfer learning can generalize. Using $\epsilon_T$ and $\epsilon^*$, we can describe these sets as neighborhoods of systems \emph{to} which we can transfer and generalize,
$$ \underbrace{\{ S_T | S_S, \epsilon_T \leq \epsilon^* \}}_{\text{Neighborhood of Targets } S_T}$$
and neighborhoods of systems \emph{from} which we can transfer and generalize,
$$ \underbrace{\{ S_S | S_T, \epsilon_T \leq \epsilon^* \}}_{\text{Neighborhood of Sources } S_S}.$$
Noting Definition \ref{def:nt}, if $\epsilon^* = \epsilon(H_T(A_T(d_T)), f_T)$, these neighborhoods are those systems to and from which transfer is positive.

The size of these neighborhoods describes the transferability of a learning system in terms of the number of systems it can transfer to or from and generalize. To the extent that cardinality gives a good description of size\footnote{Cardinality counts arbitrarily close systems as different, and it may be preferable to define a measure of equivalence, and consider the cardinality of the neighborhoods after the equivalence relation is applied.}, transferability can be defined formally as follows.
\begin{definition}{\emph{Transferability}.} \\
    Consider a target learning system $S_T$ and a source learning system $S_S$. Given a measure of error $\epsilon_T: H_{Tr}(\Theta_{Tr}) \times f_T \to \mathbb{R}$ and a threshold on error $\epsilon^* \in \mathbb{R}$, the transferability of a source is the cardinality of the neighborhood of target systems $S_T$ to which it can transfer and generalize,
    $$|\{S_T| S_S, \epsilon_T \leq \epsilon^*\}|,$$
    and the transferability of a target is the cardinality of the neighborhood of source systems $S_S$ from which we can transfer and generalize,
    $$|\{S_S|S_T, \epsilon_T \leq \epsilon^*\}|.$$
    These cardinalities are termed the source-transferability and target-transferability, respectively.
\end{definition} 
\noindent Note, this defines transferability as an attribute of a particular system---not an attribute of a source-target pairing.

Our interest in transferability as an aim of transfer learning systems echoes a growing interest of the machine learning community in a notion of \emph{generalist} learning systems \cite{kolesnikov2019big, huang2019gpipe, tschannen2020self}. Put informally, generalists are learning systems which can generalize to many tasks with few samples. Using our formalism, these systems can be described as learning systems with high source-transferability. More particularly, they can be defined as follows. 
\begin{definition}{\emph{Generalist Learning Systems}.} \\
    A generalist learning system $S_S$ is a system that can transfer to at least $t$ target systems $S_T$ with data $d_T \in D_T$ and generalize with at most $n$ target samples $(x_T, y_T) \in X_T \times Y_T.$ That is, they are systems $S_S$ where
    $$|\{S_T | S_S, |d_T| \leq n, \epsilon_T < \epsilon^*\}| \geq t$$
\end{definition}
\noindent Generalists are sources $S_S$ that can $n$-shot transfer learn to $t$ or more targets $S_T$. Generalists are typically studied in the context of deep learning for computer vision, where a single network is tasked with few-shot learning a variety of visual tasks, e.g., classification, object detection, and segmentation, in a variety of environments \cite{kolesnikov2019big}.

In the following, we go beyond existing frameworks to explore notions of transferability---and thereby generalization, transfer roughness, and transfer distance in the context of structure and behavior. In doing so, we demonstrate the mathematical depth of Definition \ref{def:tl}. We show that not only does it allow for immediate, formal consideration of surface-level phenomena covered by existing frameworks, but moreover, it allows for a considerable amount of modeling to be done at the general level, i.e., without reference to solution methods, in following with the spirit of AST depicted in Figure \ref{fig:ast-block}.

\section{Structure and Behavior in Transfer Learning}

To the extent that generalization in transfer learning is concerned with sets of systems, it is concerned with how those sets can be expressed in terms of those systems' structures and behaviors. In the following subsections, we discuss how structural and behavioral equality and, moreover, similarity relate to the difficulty of transfer learning. Equalities between $S_S$ and $S_T$ give a basic sense of the setting and what solution methods are available. Similarities between $S_S$ and $S_T$ are a richer means for elaboration, and can give a sense of the likelihood of generalization.

Learning systems are concerned with estimating functions $f:X \to Y$. As transfer learning is concerned with sharing knowledge used to estimate a source function $f_S:X_S \to Y_S$ to help estimate a target function $f_T:X_T \to Y_T$, naturally, the input-output spaces of the source $X_S \times Y_S$ and target $X_T \times Y_T$ are the principal interest of structural considerations. Similarly, the principal interest of behavioral considerations are the probability measures which specify $f_S$ and $f_T$, and, correspondingly, $D_S$ and $D_T$.

\subsection{Structural Considerations}

For source and target systems $S_S$ and $S_T$ we have the following possible equalities between system structures:
\begin{align*}
    X_S = X_T, Y_S = Y_T, \\
    X_S \neq X_T, Y_S = Y_T, \\
    X_S = X_T, Y_S \neq Y_T, \\
    X_S \neq X_T, Y_S \neq Y_T.
\end{align*}
The first case $X_S \times Y_S = X_T \times Y_T$ specifies transfer as homogeneous---all others specify heterogeneous transfer. This is the extent of discussion of structure in the existing frameworks \cite{pan2009survey, weiss2016survey}. We elaborate further.

To do so, we extend past structural equality to notions of structural similarity. Recall, structural similarity is a question of the structural homomorphism between two systems. As is common in category theory, we define a morphism as simply a map between systems, and define an onto map between systems as a homomorphism. We can investigate homomorphism in reference to a morphism $m: S_S \to S_T$. First, note that we can quantify structural similarity using equivalence classes. Let $m_x:X_S \to X_T$ and $m_y:Y_S \to Y_T$ such that $m \leftrightarrow (m_x, m_y)$. And let $S_S/m$, $X_S/m_x$, and $Y_S/m_y$ be the equivalence classes of $S_S$, $X_S$, and $Y_S$ with respect to $m$, $m_x$, and $m_y$, respectively.

\begin{figure}
    \centering
    \includegraphics{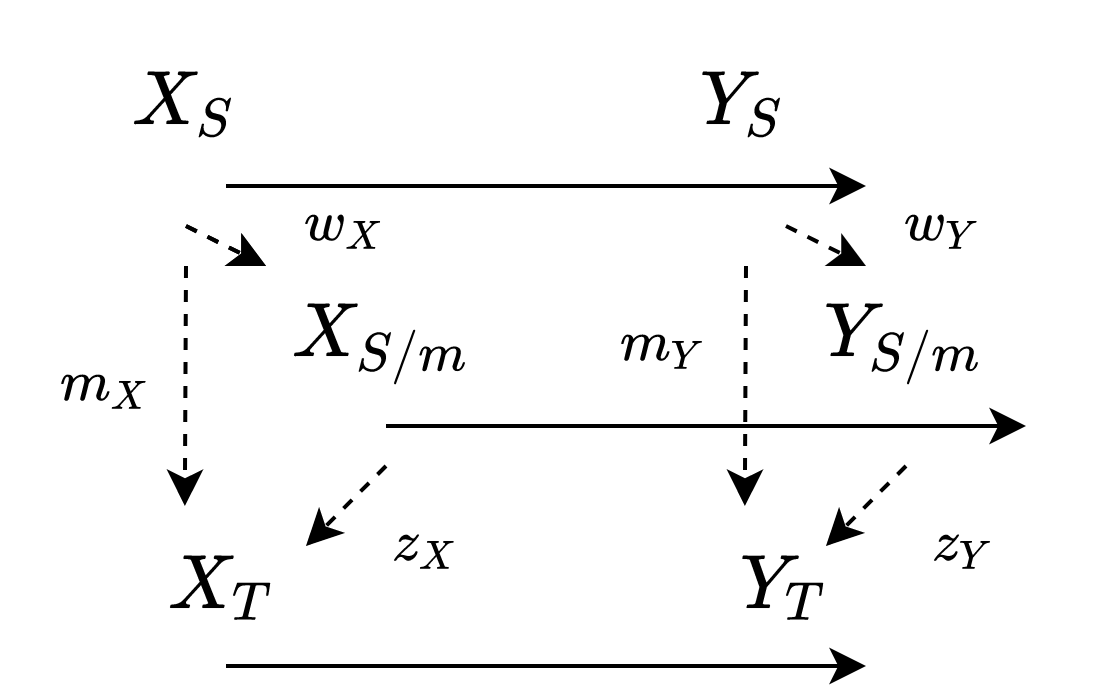}
    \caption{A commutative diagram depicting how equivalence classes can describe roughness.}
    \label{fig:roughness}
\end{figure}

Consider the two sets of relations
\begin{equation*}
  \begin{split}
    w &: S_S \to S_S/m \\
    w_x &:X_S \to X_S/m_x \\
    w_y &: Y_S \to Y_S/m_y
  \end{split}
  \qquad
  \begin{split}
    z &: S_S/m \to S_T \\
    z_x &: X_S/m \to X_T \\
    z_y &: Y_S/m \to Y_T
  \end{split}
\end{equation*}
Relation $w$ maps the source $S_S$ to its equivalence class $S_S/m$ and relation $z$ maps $S_S/m$ to the target $S_T$, as depicted by the commutative diagram shown in Figure \ref{fig:roughness}. That is,
$$S_S \xrightarrow[(w_x, w_y)]{} S_S/m \xrightarrow[(z_x, z_y)]{} S_T $$
The equivalence class $S_S/m$ describes the `roughness' of the structural similarity from $S_S$ to $S_T$. Its cardinality quantifies the `surjective-ness' of $m:S_S \to S_T$. The greater the difference between $|S_S|$ and $|S_S/m|$, the more structurally dissimilar $S_S$ and $S_T$ are. However, in the large, structural similarity is not measurable in the same way as behavioral similarity.

The homomorphism between $S_S$ and $S_T$ is better investigated in terms of the properties of $m$, such as whether it is injective, surjective, invertible, etc. For example, partial morphisms from $X_S \times Y_S$ to $X_T \times Y_T$ are associated with partial transfer \cite{cao2018partial}. When the partial morphism is surjective, only a subset of the source is transferred to the target. When the partial morphism is injective, the source transfers to only a subset of the target. Also, structural similarity can be expressed using category theory, where the structural similarity between two systems can be studied with respect to the categories of systems to which they belong. To describe structural similarity in a broad sense, we define \emph{transfer roughness} as follows.

\begin{definition}{\emph{Transfer Roughness}.} \\
    Transfer roughness describes the structural homomorphism from the source system $S_S$ to the target system $S_T$. When $S_S$ and $S_T$ are isomorphic, transfer roughness is minimal or otherwise non-existent. When roughness exists, it is defined by its properties, and thus there is no clear notion of maximal roughness.  
\end{definition}

The structure of the source relative to that of the target determines the roughness of transfer. Structures can be too dissimilar to transfer no matter what the behavior. Homomorphisms are onto and thus structure preserving, and, as such, it is a reasonable principle to characterize structural transferability in terms of the set of homomorphisms shared between the source and target. The supporting intuition is that either the source must map onto the target or they must both map onto some shared latent system, if not fully, at least in some aspect. Otherwise information in the source is lost when transferring to the target.

Let $\mathcal{H}(X, Y)$ denote the set of all structures homomorphic to $X \times Y$. The set of homomorphic structures between $S_S$ and $S_T$ is given by,
$$\mathcal{H}(X_S, Y_S) \cap \mathcal{H}(X_T, Y_T).$$
In transfer learning, we are specifically interested in using knowledge from $S_S$ to help learn $f_T$. Thus, not all elements of this intersection are valid structures for transfer learning, only those whose output can be mapped to $Y_T$. This set of valid structures can be expressed as,
$$\mathcal{V} = \{X \times Y \in \mathcal{H}(X_S, Y_S) \cap \mathcal{H}(X_T, Y_T) | \exists m_y:Y \to Y_T\}.$$
Apparently not all elements of $\mathcal{V}$ will be useful structures for estimating $f_T$, however, those that are useful, presuming structural homomorphism is necessary, will be in $\mathcal{V}$.

If we define $\mathcal{V}'$ to be the subset of $\mathcal{V}$ where transfer learning generalizes, i.e., the homomorphic structures where $\epsilon_T < \epsilon^*$, transferability can be defined in structural terms as follows.
\begin{definition}{\emph{Structural Transferability}.} \\
    Consider a target learning system $S_T$ and a source learning system $S_S$. The structural transferability of a source $S_S$ is,
    $$|\{S_T|S_S, \exists (X \times Y) \in \mathcal{V}'(S_S, S_T)\}|,$$
    and the structural transferability of a target is,
    $$|\{S_S|S_T, \exists (X \times Y) \in \mathcal{V}'(S_S, S_T)\}|.$$
\end{definition}
\noindent In other words, structural transferability concerns the set of systems that share a useful homomorphism with $S_S$ and $S_T$. While in practice $\mathcal{V}$ and $\mathcal{V}'$ are difficult to determine, they provide a theoretical basis for considering whether transfer learning is structurally possible between two systems and the structural invariance of the usefulness of transferred knowledge, respectively.

The relation $\mathcal{V}' \subset \mathcal{V}$ is particularly difficult. Ordering structural usefulness by homomorphism alone is difficult because of the vagueness of how homomorphism can be measured. The more isomorphism there is between $S_S$ and $S_T$, the more the question of usefulness shifts to the behavior. There, the error $\epsilon$ provides the ordering\footnote{$\epsilon$ is a transfer distance between posteriors specifying $h(\theta)$ and $f$.} and the threshold $\epsilon^*$ provides the partition. Structural similarity provides no clear parallel. 

It is true that if no homomorphism exists between $S_S$ and $S_T$, they are from different categories. While functors can be used to map between categories, they necessarily distort transferred knowledge because they must add or remove structure to do so. Homomorphisms between systems, in contrast, are structure preserving. And so perhaps a partial order between homomorphic and non-homomorphic systems is justified. But this ordering is hardly granular. A more formal digression on this topic is beyond the scope of this paper, but well within the scope of AST\cite{mesarovic1989abstract}.

\begin{exmp}{\emph{Transfer Roughness in UAVs.}} \\
    Consider $S_S$, $S_T$, and $S_{Tr}$ defined according to Example \ref{ex:transfer}. From Example \ref{ex:learning-systems} $X_S \times Y_S = X_T \times Y_T$, so $S_{Tr}$ involves homogeneous transfer. But, if $X_T$ did not include radar, transfer would be heterogeneous. Similarly so if $Y_S$ described paths up to 100 meters in length and $Y_T$ paths up to 10 meters. In either case, $X_S \times Y_S$ can map onto $X_T \times Y_T$, but $X_T \times Y_T$ cannot map onto $X_S \times Y_T$. Thus, transfer from $S_T$ to $S_S$ is rougher than transfer from $S_S$ to $S_T$.
\end{exmp}

\subsection{Behavioral Considerations}

In transfer learning, the primary behaviors of interest are $P(X)$ and $P(Y|X)$ from the domain $\mathcal{D}$ and task $\mathcal{T}$, respectively, and the joint distribution they form,
$$P(X,Y) = P(X) P(Y|X).$$ 
It is important to realize that $P(X_S, Y_S) \neq P(X_T, Y_T)$ only implies that $P(X_S) \neq P(X_T) \lor P(Y_S|X_S) \neq P(Y_T|X_T)$. That is, the posteriors $P(Y|X)$ can still be equal when the joints $P(X, Y)$ are not if the marginals $P(X)$ offset the difference, and vice versa. In the main, these behavioral equalities make absolute statements on the inductive or transductive nature of a transfer learning system. Behavioral similarities, in contrast, have the richness to make statements on the likelihood of generalization, and, thereby, on transferability.

In AST, behavior is a topological-type concept and, accordingly, behavioral similarity is akin to a generalized metric. However, because in transfer learning we are concerned primarily with behaviors which are probability measures, behavioral similarity between $S_S$ and $S_T$ takes the form of distributional divergences. In our elaboration of behavioral similarity we focus on a notion of \emph{transfer distance}. Transfer distance is the abstract distance knowledge must traverse to be transferred from one system to another. We consider it to be a measure on the input spaces $X_S \times X_T$, output spaces $Y_S \times Y_T$, or input-output spaces $(X_S \times Y_S) \times (X_T \times Y_T)$---more specifically, as a measure on probability measures over those spaces. It can be defined formally as follows.

\begin{definition}{\emph{Transfer Distance}.} \\
    Let $S_S$ and $S_T$ be source and target learning systems. Let $Z_i$ be a non-empty element of $\mathcal{P}(X_i \cup Y_i)$. Transfer distance $\delta_T$ is a measure
    $$\delta_T:P(Z_S) \times P(Z_T) \to \mathbb{R}$$
    of distance between the probability measures $P(Z_i)$ related to the estimated functions $f_i:X_i \to Y_i$ of $S_S$ and $S_T$.
\end{definition}

In practice, transfer distances are often given by $f$-divergences \cite{ditzler2011hellinger}, such as KL-divergence or the Hellinger distance, Wasserstein distances \cite{shen2017wasserstein}, and maximum mean discrepancy \cite{pan2008transfer, long2017deep, jiang2015integration}. Others use generative adversarial networks, a deep learning distribution modeling technique, to estimate divergence \cite{tzeng2015simultaneous, ganin2016domain}. Commonly, these distances are used to calculate divergence-based components of loss functions. Herein, we consider transfer distance's more general use in characterizing transfer learning systems. 

In heterogeneous transfer, transfer distances can be used after feature-representation transfer has given the probability measures of interest the same support. Transfer distances between measures with different support are not widely considered in existing machine learning literature. However, the assumptions of homogeneous transfer and domain adaptation, i.e., $X_S \times Y_S = X_T \times Y_T$, allow for a rich theory of the role of transfer distance in determining the upper-bound on error. 

Upper-bounds on $\epsilon_T$ have been given in terms of statistical divergence \cite{blitzer2008learning}, $H$-divergence \cite{ben2010theory}, Rademacher complexity \cite{mohri2009rademacher}, and integral probability metrics \cite{zhang2012generalization}, among others. Despite their differences, central to most is a transfer distance $\delta_T: P(X_S) \times P(X_T) \to \mathbb{R}$ that concerns the closeness of input behavior and a term $C$ that concerns the complexity of estimating $f_T$. These bounds roughly generalize to the form,
\begin{equation}
\epsilon_T \leq \epsilon_S + \delta_T + C
\label{eq:inequality}
\end{equation}
where $\epsilon_T$ and $\epsilon_S$ are the errors in $S_T$ and $S_S$, $\delta_T$ is the transfer distance, and $C$ is a constant term. $C$ is often expressed in terms of sample sizes, e.g., $|D_S|$ and $|D_T|$, capacity, e.g., the VC-dimension of $H_T$ \cite{ben2010theory}, and information complexity, e.g., the Rademacher complexity of $D_T$ \cite{mohri2009rademacher}. Note, closeness and complexity are often not as separable as suggested by Inequality \ref{eq:inequality}. 

To the extent that Inequality \ref{eq:inequality} holds, we can describe transferability in terms of transfer distance. Generalization in transfer learning occurs if $\epsilon_T \leq \epsilon^*$, and since $\epsilon_T \leq \epsilon_S + \delta_T + C$, $\epsilon_S + \delta_T + C \leq \epsilon^* \implies \epsilon_T \leq \epsilon^*$. Thus, transferability can be defined in behavioral terms as follows.
\begin{definition}{\emph{Behavioral Transferability}.} \\
    Consider a target learning system $S_T$ and a source learning system $S_S$. The behavioral transferability of a source $S_S$ is,
    $$|\{S_T|S_S, \epsilon_S + \delta_T + C < \epsilon^*\}|,$$
    and the behavioral transferability of a target is,
    $$|\{S_S|S_T, \epsilon_S + \delta_T + C < \epsilon^*\}|.$$
\end{definition}
\noindent For $S_S$ with similar $\epsilon_S$ and $S_T$ with similar $C$, given a threshold on distance $\delta^* \in \mathbb{R}$, behavioral transferability can be expressed entirely in terms of transfer distance:
$$|\{S_T | S_S, \delta_T < \delta^*\}| \text{ and } |\{S_S | S_T, \delta_T < \delta^*\}|.$$
Of course, specific bounds on $\epsilon_T$ with specific distances $\delta_T$ from the literature can be substituted in the stead of Inequality \ref{eq:inequality}. Also note, we are assuming $X_S \times Y_S = X_T \times Y_T$. When $X_S \times Y_S \neq X_T \times Y_T$, transfer distance is a measure between probability measures with different supports, and while an upper-bound like Inequality \ref{eq:inequality} may be appropriate, it is not supported by existing literature. In such cases it is important to consider structural similarity.

\begin{exmp}{\emph{Transfer Distance in UAVs.}} \\
    Consider $S_S$, $S_T$, and $S_{Tr}$ defined according to Example \ref{ex:transfer}. Let source $S_S$ be associated with a desert biome and $S_T$ a jungle biome. When comparing $P(X_T)$ to $P(X_S)$, increased foliage in $S_T$ suggests accelerometer readings with higher variance, camera images with different hue, saturation, and luminance, and radar readings with more obstacles. Similarly, increased foliage may also mean paths in $P(Y_T|X_T)$ must compensate more for uncertainty than those in $P(Y_S|X_S)$. In contrast, foliage is more similar between the desert and tundra, thus, transfer distance is likely larger from the desert to the jungle than from the desert to the tundra.
\end{exmp}

\subsection{Remarks}

In summary, structure and behavior provide a means of elaborating deeply on transfer learning systems, just as they do for systems writ large. Structural considerations center on the structural relatability of $S_S$ and $S_T$ and the usefulness of the related structures $X \times Y$ for transfer learning. Behavioral considerations center on the behavioral closeness of $S_S$ and $S_T$ and the complexity of learning $f_T$. These concerns provide guideposts for the design and analysis of transfer learning systems. While the joint consideration of structure and behavior is necessary for a complete perspective on transfer learning systems, herein, in following with broader systems theory, we advocate that their joint consideration ought to come from viewing structure and behavior as parts of a whole---instead of approaching their joint consideration directly by neglecting notions of structure and behavior entirely, as is advocated implicitly by the existing frameworks pervasive use of domain $\mathcal{D}$ and task $\mathcal{T}$.

\section{Conclusion}

Our framework synthesizes systems theoretic notions of structure and behavior with key concepts in transfer learning. These include homogeneous and heterogeneous transfer, domain adaptation, inductive and transductive transfer, negative transfer, and more. In subsequent elaborations, we provide formal descriptions of transferability, transfer roughness, and transfer distance, all in reference to structure and behavior.

This systems perspective places emphasis on different aspects of transfer learning than existing frameworks. When we take behavior to be represented by a posterior or joint distribution, we arrive at constructs similar to existing theory. More distinctly, when we introduce structure, and study it in isolation, we arrive at notions of roughness, homomorphism, and category neglected in existing literature.

The presented framework offers a formal approach for modeling learning. The focal points of our theory are in aspects central to the general characterization and categorization of transfer learning as a mathematical construct, not aspects central to scholarship. This strengthens the literature by contributing a framework that is more closely rooted to engineering design and analysis than existing frameworks. Because our framework is pointedly anchored to concepts from existing surveys, practitioners should face little difficulty in the simultaneous use of both. Taken together, practitioners have a modeling framework and a reference guide to the literature.

Herein, we have modeled transfer learning as a subsystem. Transfer learning systems can be connected component-wise to the systems within which they are embedded. Subsequently, deductions can be made regarding the design and operation of systems and their learning subsystems with the interrelationships between them taken into account. In this way, we contribute a formal systems theory of transfer learning to the growing body of engineering-centric frameworks for machine learning.

Real-world systems need transfer learning, and, correspondingly, engineering frameworks to guide its application. The presented framework offers a Mesarovician foundation.

\section{Appendix}

\subsection{Mesarovician Glossary}

\begin{definition}{\emph{System Behavior}.} \\
    System behaviors are properties or descriptions paired with systems. For example, consider a system $S:X \to Y$ and a map $S \to \{ stable, neutral, unstable \}$ or from $S \to P(X,Y)$. System behavior is a topological-type concept in the sense that it pairs systems with elements of sets of behaviors.
\end{definition}

\begin{definition}{\emph{Behavioral Similarity}.} \\
    Behavioral similarity describes the `proximity' between two systems' behavior. To the extent that behavior can be described topologically, behavioral similarity can be expressed in terms of generalized metrics (topological `distance'), metrics and pseudo-metrics (measure theoretic `distance'), and statistical divergences (probability/information theoretic `distance'), depending on the nature of the topology.
\end{definition}

\begin{definition}{\emph{System Structure}.} \\
    System structure is the mathematical structure of a system's component sets and the relations among them. For example, there may be algebraic structure, e.g. the linearity of a relationship between two component sets, related to the definition of the relation.
\end{definition}

\begin{definition}{\emph{Structural Similarity}.} \\
    Structural similarity describes the homomorphism between two systems' structures. It is described in reference to a relation $m:S_1 \to S_2$, termed a morphism. The equivalence class $S_1/m$ describes the `roughness' of the structural similarity between $S_1$ and $S_2$. Its cardinality gives a quantity to the `surjective-ness' of $m:S_1 \to S_2$. However, in the large, structural similarity is not measurable in the same way as behavioral similarity. The homomorphism is better studied using properties of $m$.
\end{definition}

\begin{definition}{\emph{Cascade Connection}.} \\
    Let $\circ: \overline{S} \times \overline{S} \to \overline{S}$ be such that $S_1 \circ S_2 = S_3$, where,
    \begin{gather*}
        S_1 \subset X_1 \times (Y_1 \times (Z_1)), S_2 \subset (X_2 \times Z_2) \times Y_2 \\
        S_3 \subset (X_1 \times X_2) \times (Y_1 \times Y_2), Z_1 = Z_2 = Z
    \end{gather*}
    and,
    \begin{gather*}
        ((x_1, x_2), (y_1, y_2)) \in S_3 \leftrightarrow \\
        (\exists z) ((x_1, (y_1, z)) \in S_1 \wedge ((x_2, z), y_2) \in S_2)
    \end{gather*}
    $\circ$ is termed the cascade (connecting) operator.
\end{definition}

\subsection{Learning Systems}

\begin{prop}
    \emph{$S$ in Definition \ref{def:ls} is a cascade connection of two input-output systems.} \\
    \emph{Proof:}
    Recall $S \subset \times \{A, D, \Theta, H, X, Y\}$. First we will show $A$ and $H$ to be input-output systems. First note that $A \subset \times \{ D, \Theta \}$. Noting $D \subset X \times Y$, apparently $D \cap \Theta = \emptyset$ and $D \cup \Theta = \overline{A}$. Similarly, $H \subset \times \{ \Theta, X, Y \}$. Letting $X' = \{ X, \Theta \}$, apparently $X' \cap Y = \emptyset$ and $X' \cup Y = \overline{H}$. Therefore, by definition, $A$ and $H$ are input-output systems. Let $S_C:D \times X \to Y$. Apparently, for $d \in D, x \in X, y \in Y, \theta \in \Theta$, $((d, x), y) \in S_C \leftrightarrow \exists \theta ((d, \theta) \in A \wedge (\theta, x, y) \in H$. Therefore, $S_C: A \circ H$. Lastly, note $S_C$ is a function-type representation of $S$, where $A$, $H$, and $\Theta$ are left as specifications on relations, not included as component sets.
\end{prop}

\begin{prop}
    \emph{$S$ in Definition \ref{def:gsls} is a goal-seeking system.} \\
    \emph{Proof:}
    Goal-seeking is characterized by the consistency relations $(G, E)$ and by the internal feedback of $X \times Y$ into $S_G$. Note $D \subset X \times Y$ satisfies internal feedback. The consistency relations $(G, E)$ in Definition \ref{def:gs} and \ref{def:gsls} can be shown to be isomorphic by substituting $D \subset X \times Y$ into consistency relations $G$ and $E$ in Definition \ref{def:gs} and $(x, y) \in d$ into their constraints. Thus, by definition, $S$ in Definition \ref{def:gsls} is a goal-seeking system, where $S_G$ is the inductive system $A$ and $S_F$ is the functional system $H$. 
\end{prop}

\begin{prop}
    \emph{Empirical risk minimization is a special case of a learning system as defined in Definition \ref{def:gsls}.} \\
    \emph{Proof:}
    A learning system given by Definition \ref{def:gsls} is an empirical risk minimization learning system if (1) $D$ is a sample of $l$ independent and identically distributed observations sampled according to an unknown distribution $P(X, Y)$, and (2) $A$ selects $\theta \in \Theta$ by minimizing the empirical risk $R_{emp}$, calculated on the basis of $D$, over $\theta \in \Theta$. Otherwise put, ERM is a learning system $S \subset \times \{A, D, \Theta, G, E, H, X, Y\}$ where $G(D, \theta) = R_{emp}(D, \theta) = \frac{1}{l}\sum\limits_{i=1}^l L(y_i, h(x_i, \theta))$ and $E = \min_{\theta \in \Theta} G(D, \theta)$, where $L$ is a loss function. 
\end{prop}

\bibliographystyle{IEEEtran}
\bibliography{ref}

\end{document}